\documentclass[final]{cvpr}
\usepackage{nopageno}

\usepackage{times}
\usepackage{epsfig}
\usepackage{graphicx}
\usepackage{amsmath}
\usepackage{amssymb}

\usepackage{caption}
\usepackage{subcaption}
\usepackage{booktabs}
\usepackage{tabu}
\usepackage{multirow}
\usepackage{pifont}
\usepackage{color}
\usepackage[numbers,sort&compress]{natbib}

\newcommand{\cmark}{\text{\ding{51}}}

\usepackage[pagebackref=true,breaklinks=true,colorlinks,bookmarks=false]{hyperref}

\def\cvprPaperID{5077} 
\def\confYear{CVPR 2021}

\begin{document}

\title{Progressive Semantic Segmentation}


\author{
Chuong Huynh$^{1}$ \quad Anh Tuan Tran$^{1,2}$ \quad Khoa Luu$^{1,3}$ \quad Minh Hoai$^{1,4}$ \\
$^1$VinAI Research, Hanoi, Vietnam,  $^2$VinUniversity, Hanoi, Vietnam\\
$^3$University of Arkansas, Fayetteville, AR 72701, USA\\
$^4$Stony Brook University, Stony Brook, NY 11790, USA\\
{\tt\small \{v.chuonghm,v.anhtt152,v.khoal,v.hoainm\}@vinai.io}
}

\maketitle

\newcommand{\mh}[1]{\textcolor{blue}{[Hoai: {#1}]}}
\newcommand{\anh}[1]{{\textcolor{cyan}{[Anh: #1]}}}
\newcommand{\del}[1]{\textcolor{purple}{[Remove: {#1}]}}
\newcommand{\add}[1]{\textcolor{red}{#1}}
\newcommand{\Sref}[1]{Sec.~\ref{#1}}
\newcommand{\Eref}[1]{Eq.~(\ref{#1})}
\newcommand{\Fref}[1]{Fig.~\ref{#1}}
\newcommand{\Tref}[1]{Table~\ref{#1}}

\def\mA{\mathcal{A}}
\def\mB{\mathcal{B}}
\def\mC{\mathcal{C}}
\def\mD{\mathcal{D}}
\def\mE{\mathcal{E}}
\def\mF{\mathcal{F}}
\def\mG{\mathcal{G}}
\def\mH{\mathcal{H}}
\def\mI{\mathcal{I}}
\def\mJ{\mathcal{J}}
\def\mK{\mathcal{K}}
\def\mL{\mathcal{L}}
\def\mM{\mathcal{M}}
\def\mN{\mathcal{N}}
\def\mO{\mathcal{O}}
\def\mP{\mathcal{P}}
\def\mQ{\mathcal{Q}}
\def\mR{\mathcal{R}}
\def\mS{\mathcal{S}}
\def\mT{\mathcal{T}}
\def\mU{\mathcal{U}}
\def\mV{\mathcal{V}}
\def\mW{\mathcal{W}}
\def\mX{\mathcal{X}}
\def\mY{\mathcal{Y}}
\def\mZ{\mathcal{Z}} 

\def\bbN{\mathbb{N}} 
\def\bbR{\mathbb{R}} 
\def\bbP{\mathbb{P}} 
\def\bbQ{\mathbb{Q}} 
\def\bbE{\mathbb{E}}

\def\1n{\mathbf{1}_n}
\def\0{\mathbf{0}}
\def\1{\mathbf{1}}

\def\A{{\bf A}}
\def\B{{\bf B}}
\def\C{{\bf C}}
\def\D{{\bf D}}
\def\E{{\bf E}}
\def\F{{\bf F}}
\def\G{{\bf G}}
\def\H{{\bf H}}
\def\I{{\bf I}}
\def\J{{\bf J}}
\def\K{{\bf K}}
\def\L{{\bf L}}
\def\M{{\bf M}}
\def\N{{\bf N}}
\def\O{{\bf O}}
\def\P{{\bf P}}
\def\Q{{\bf Q}}
\def\R{{\bf R}}
\def\S{{\bf S}}
\def\T{{\bf T}}
\def\U{{\bf U}}
\def\V{{\bf V}}
\def\W{{\bf W}}
\def\X{{\bf X}}
\def\Y{{\bf Y}}
\def\Z{{\bf Z}}

\def\a{{\bf a}}
\def\b{{\bf b}}
\def\c{{\bf c}}
\def\d{{\bf d}}
\def\e{{\bf e}}
\def\f{{\bf f}}
\def\g{{\bf g}}
\def\h{{\bf h}}
\def\i{{\bf i}}
\def\j{{\bf j}}
\def\k{{\bf k}}
\def\l{{\bf l}}
\def\m{{\bf m}}
\def\n{{\bf n}}
\def\o{{\bf o}}
\def\p{{\bf p}}
\def\q{{\bf q}}
\def\r{{\bf r}}
\def\s{{\bf s}}
\def\t{{\bf t}}
\def\u{{\bf u}}
\def\v{{\bf v}}
\def\w{{\bf w}}
\def\x{{\bf x}}
\def\y{{\bf y}}
\def\z{{\bf z}}

\def\balpha{\mbox{\boldmath{$\alpha$}}}
\def\bbeta{\mbox{\boldmath{$\beta$}}}
\def\bdelta{\mbox{\boldmath{$\delta$}}}
\def\bgamma{\mbox{\boldmath{$\gamma$}}}
\def\blambda{\mbox{\boldmath{$\lambda$}}}
\def\bsigma{\mbox{\boldmath{$\sigma$}}}
\def\btheta{\mbox{\boldmath{$\theta$}}}
\def\bomega{\mbox{\boldmath{$\omega$}}}
\def\bxi{\mbox{\boldmath{$\xi$}}}
\def\bnu{\mbox{\boldmath{$\nu$}}}                                  
\def\bphi{\mbox{\boldmath{$\phi$}}}
\def\bmu{\mbox{\boldmath{$\mu$}}}

\def\bDelta{\mbox{\boldmath{$\Delta$}}}
\def\bOmega{\mbox{\boldmath{$\Omega$}}}
\def\bPhi{\mbox{\boldmath{$\Phi$}}}
\def\bLambda{\mbox{\boldmath{$\Lambda$}}}
\def\bSigma{\mbox{\boldmath{$\Sigma$}}}
\def\bGamma{\mbox{\boldmath{$\Gamma$}}}
                                  
\newcommand{\myprob}[1]{\mathop{\mathbb{P}}_{#1}}

\newcommand{\myexp}[1]{\mathop{\mathbb{E}}_{#1}}

\newcommand{\mydelta}[1]{1_{#1}}

\newcommand{\myminimum}[1]{\mathop{\textrm{minimum}}_{#1}}
\newcommand{\mymaximum}[1]{\mathop{\textrm{maximum}}_{#1}}    
\newcommand{\mymin}[1]{\mathop{\textrm{minimize}}_{#1}}
\newcommand{\mymax}[1]{\mathop{\textrm{maximize}}_{#1}}
\newcommand{\mymins}[1]{\mathop{\textrm{min.}}_{#1}}
\newcommand{\mymaxs}[1]{\mathop{\textrm{max.}}_{#1}}  
\newcommand{\myargmin}[1]{\mathop{\textrm{argmin}}_{#1}} 
\newcommand{\myargmax}[1]{\mathop{\textrm{argmax}}_{#1}} 
\newcommand{\myst}{\textrm{s.t. }}

\newcommand{\denselist}{\itemsep -1pt}
\newcommand{\sparselist}{\itemsep 1pt}

\definecolor{pink}{rgb}{0.9,0.5,0.5}
\definecolor{purple}{rgb}{0.5, 0.4, 0.8}   
\definecolor{gray}{rgb}{0.3, 0.3, 0.3}
\definecolor{mygreen}{rgb}{0.2, 0.6, 0.2}

\newcommand{\cyan}[1]{\textcolor{cyan}{#1}}
\newcommand{\red}[1]{\textcolor{red}{#1}}  
\newcommand{\blue}[1]{\textcolor{blue}{#1}}
\newcommand{\magenta}[1]{\textcolor{magenta}{#1}}
\newcommand{\pink}[1]{\textcolor{pink}{#1}}
\newcommand{\green}[1]{\textcolor{green}{#1}} 
\newcommand{\gray}[1]{\textcolor{gray}{#1}}    
\newcommand{\mygreen}[1]{\textcolor{mygreen}{#1}}    
\newcommand{\purple}[1]{\textcolor{purple}{#1}}       

\definecolor{greena}{rgb}{0.4, 0.5, 0.1}
\newcommand{\greena}[1]{\textcolor{greena}{#1}}

\definecolor{bluea}{rgb}{0, 0.4, 0.6}
\newcommand{\bluea}[1]{\textcolor{bluea}{#1}}
\definecolor{reda}{rgb}{0.6, 0.2, 0.1}
\newcommand{\reda}[1]{\textcolor{reda}{#1}}

\def\changemargin#1#2{\list{}{\rightmargin#2\leftmargin#1}\item[]}
\let\endchangemargin=\endlist
                                               
\newcommand{\cm}[1]{}

\newcommand{\mhoai}[1]{{\color{blue}\textbf{[Hoai: #1]}}}

\newcommand{\mtodo}[1]{{\color{red}$\blacksquare$\textbf{[TODO: #1]}}}
\newcommand{\myheading}[1]{\vspace{1ex}\noindent \textbf{#1}}
\newcommand{\htimesw}[2]{\mbox{$#1$$\times$$#2$}}


\newif\ifshowsolution
\showsolutiontrue

\ifshowsolution  
\newcommand{\Comment}[1]{\paragraph{\bf $\bigstar $ COMMENT:} {\sf #1} \bigskip}
\newcommand{\Solution}[2]{\paragraph{\bf $\bigstar $ SOLUTION:} {\sf #2} }
\newcommand{\Mistake}[2]{\paragraph{\bf $\blacksquare$ COMMON MISTAKE #1:} {\sf #2} \bigskip}
\else
\newcommand{\Solution}[2]{\vspace{#1}}
\fi

\newcommand{\truefalse}{
\begin{enumerate}
	\item True
	\item False
\end{enumerate}
}

\newcommand{\yesno}{
\begin{enumerate}
	\item Yes
	\item No
\end{enumerate}
}

\begin{abstract}
The objective of this work is to segment high-resolution images without overloading GPU memory usage or losing the fine details in the output segmentation map. The memory constraint means that we must either downsample the big image or divide the image into local patches for separate processing. However, the former approach would lose the fine details, while the latter can be ambiguous due to the lack of a global picture. 
In this work, we present MagNet, a multi-scale framework that resolves local ambiguity by looking at the image at multiple magnification levels. MagNet has multiple processing stages, where each stage corresponds to a magnification level, and the output of one stage is fed into the next stage for coarse-to-fine information propagation. Each stage analyzes the image at a higher resolution than the previous stage, recovering the previously lost details due to the lossy downsampling step, and the segmentation output is progressively refined through the processing stages. Experiments on three high-resolution datasets of urban views, aerial scenes, and medical images show that MagNet consistently outperforms the state-of-the-art methods by a significant margin. Code is available at \url{https://github.com/VinAIResearch/MagNet}.

   
\end{abstract}
\vspace{-1.0em}

\section{Introduction}
\vspace{-0.25em}


\newcommand{\subfigwidth}{0.49\columnwidth}
\newcommand{\heightwidth}{0.4\columnwidth}
\begin{figure}
    \centering
    \begin{subfigure}[b]{\subfigwidth}
        \centering
        \makebox[\subfigwidth]{\small{(a) Input image}}
    	\includegraphics[width=\columnwidth, height=\heightwidth]{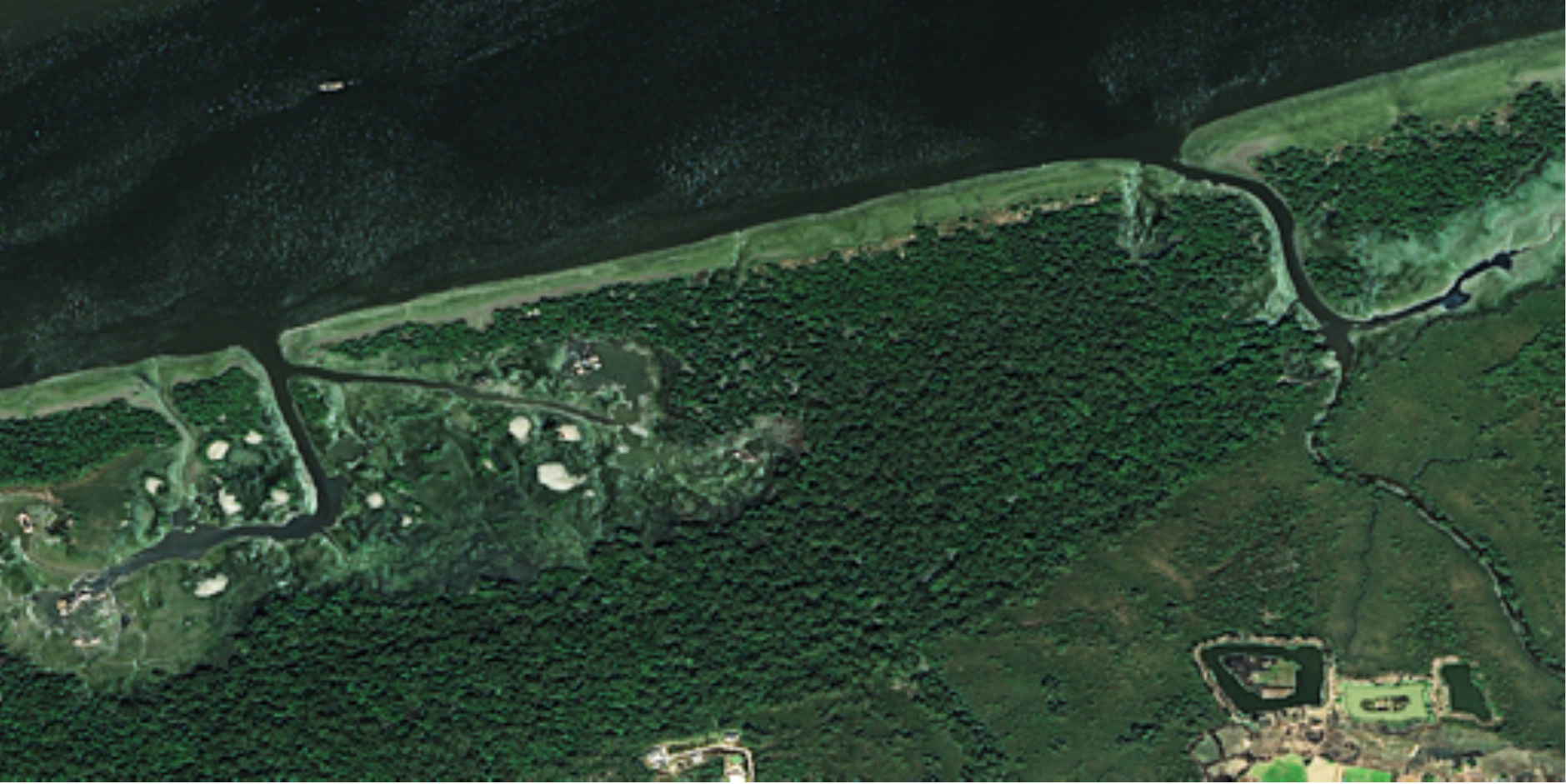}
    \end{subfigure} 
    \begin{subfigure}[b]{\subfigwidth}
        \centering
        \makebox[\subfigwidth]{\small{(b) Ground truth}}
    	\includegraphics[width=\columnwidth, height=\heightwidth]{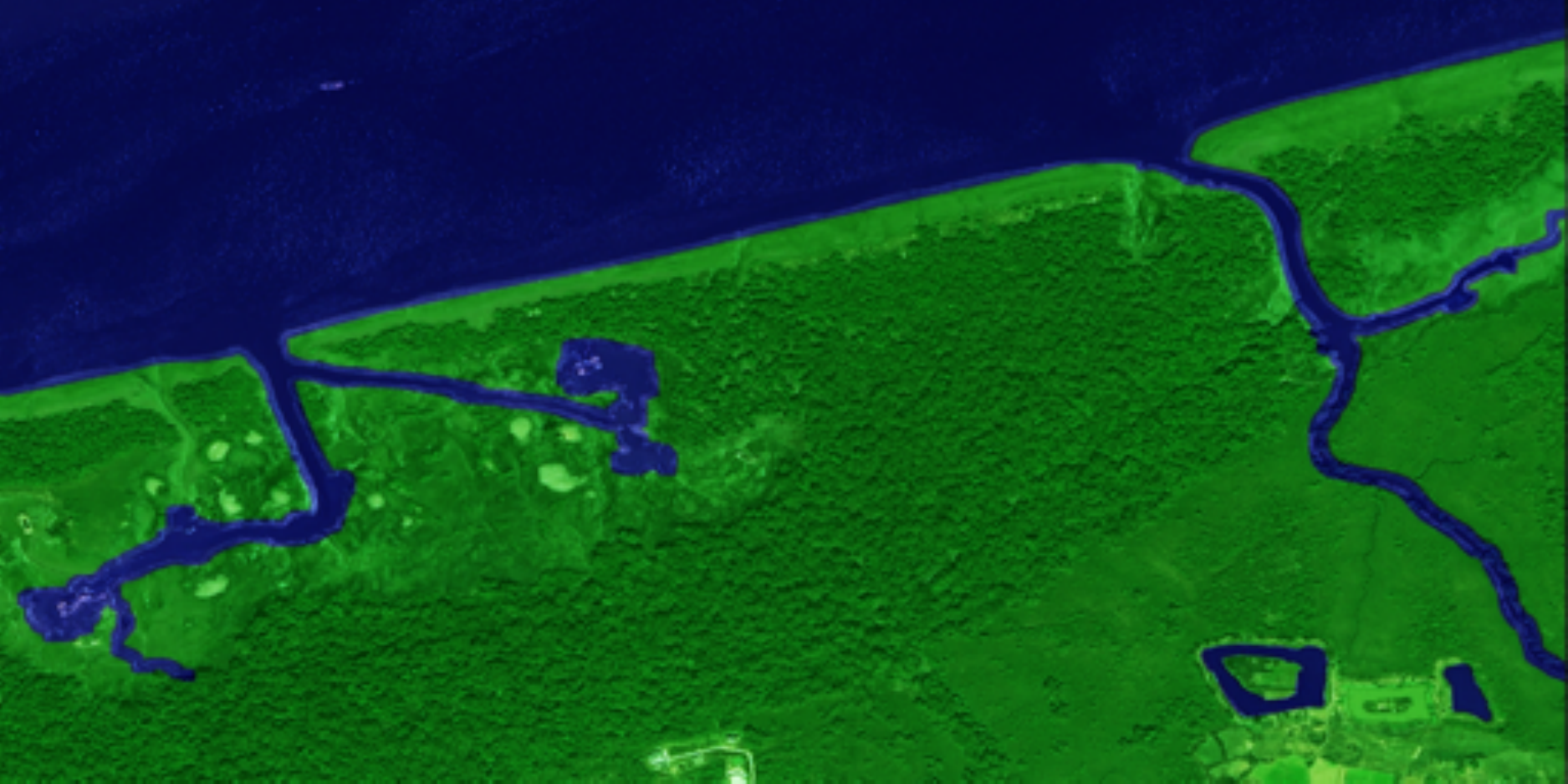}
    \end{subfigure}
    \vspace*{2mm}

    \begin{subfigure}[b]{\subfigwidth}
        \centering
        \makebox[\subfigwidth]{\small{(c) Downsampling}}
    	\includegraphics[width=\columnwidth, height=\heightwidth]{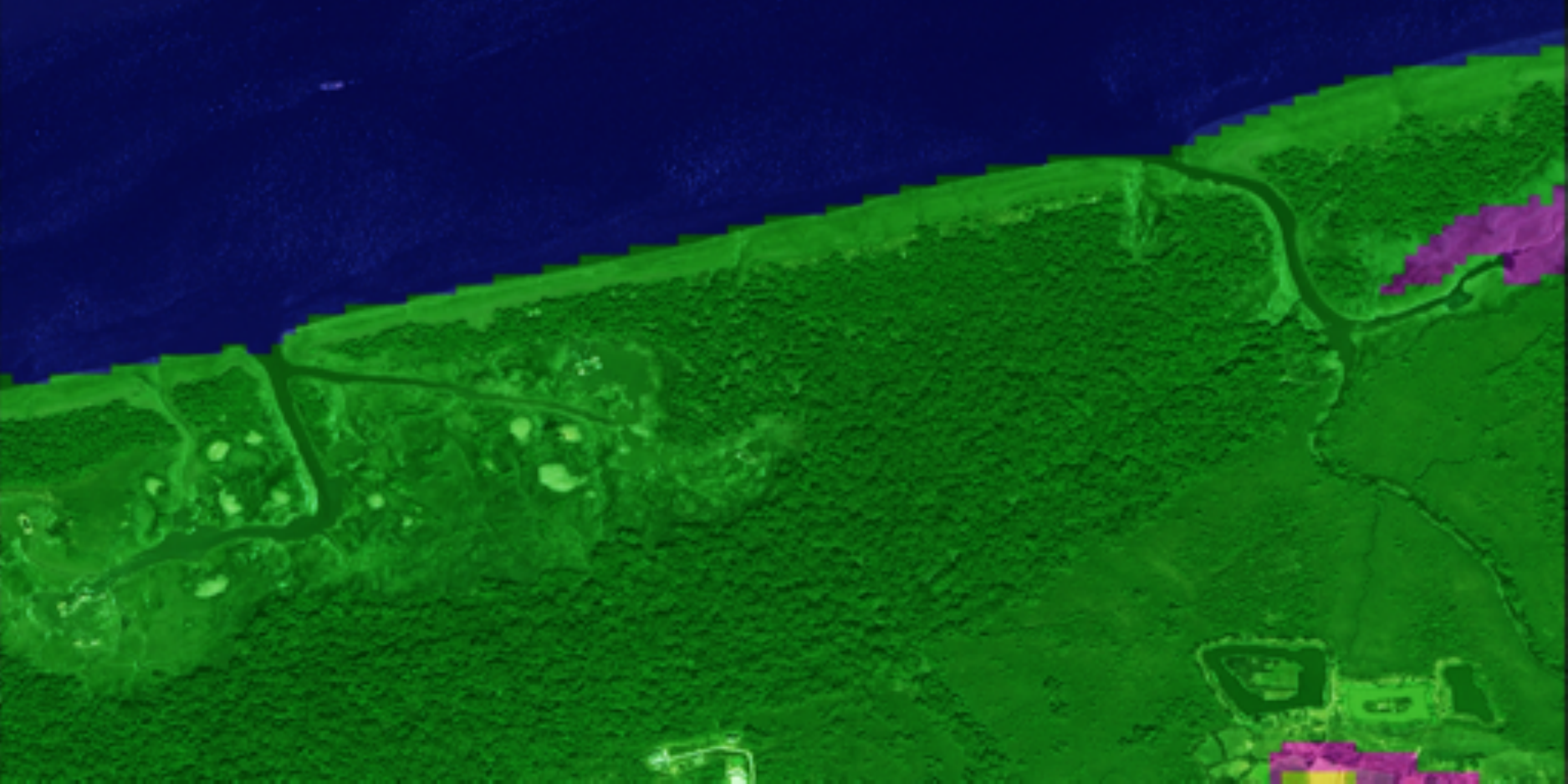}
    \end{subfigure}
    \begin{subfigure}[b]{\subfigwidth}
        \centering
        \makebox[\subfigwidth]{\small{(d) Patch processing}}
    	\includegraphics[width=\columnwidth, height=\heightwidth]{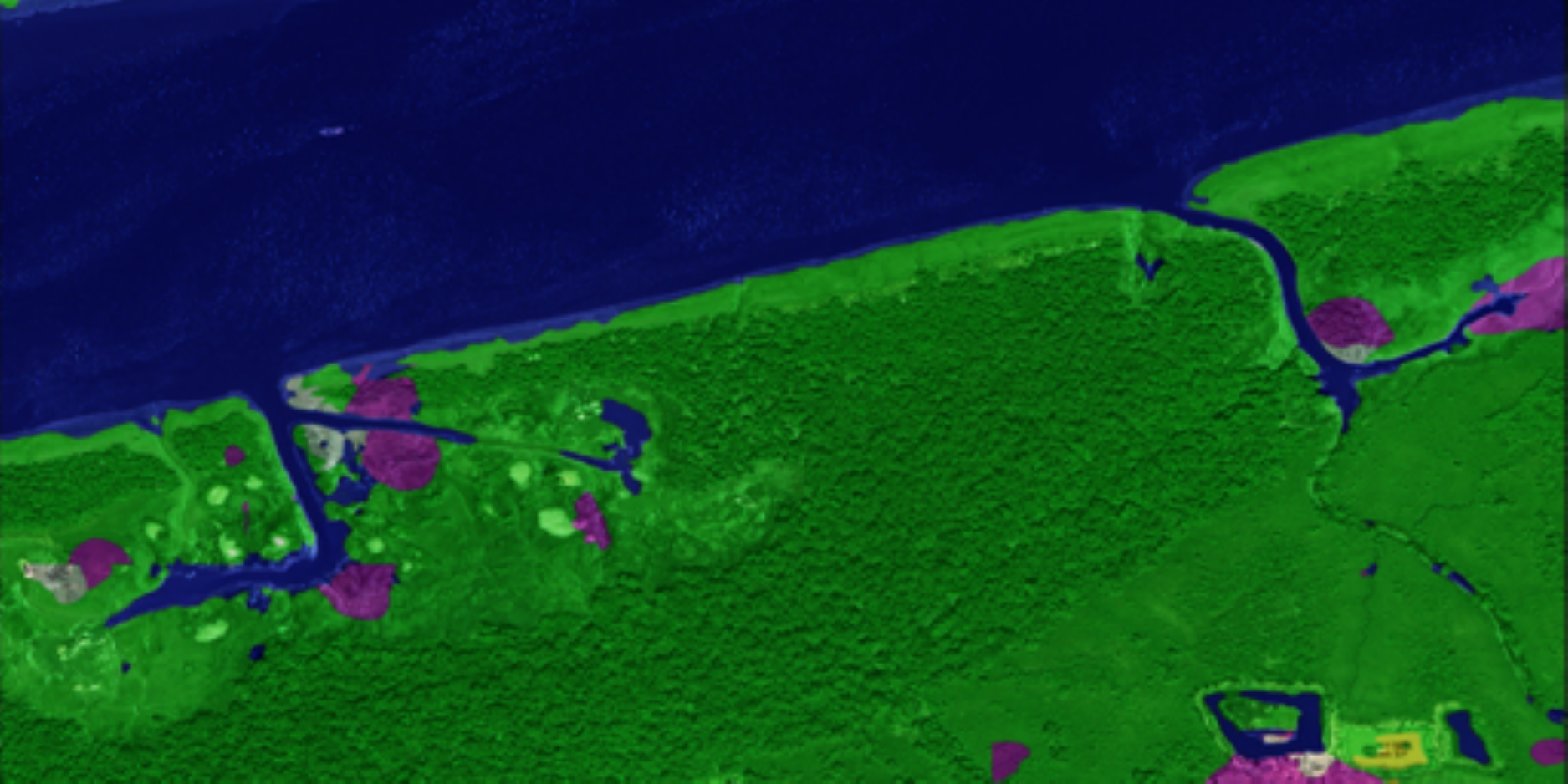}
    \end{subfigure}
    \vspace*{2mm}
    
    \begin{subfigure}[b]{\subfigwidth}
        \centering
        \makebox[\subfigwidth]{\small{(e) GLNet}}
    	\includegraphics[width=\columnwidth,height=\heightwidth]{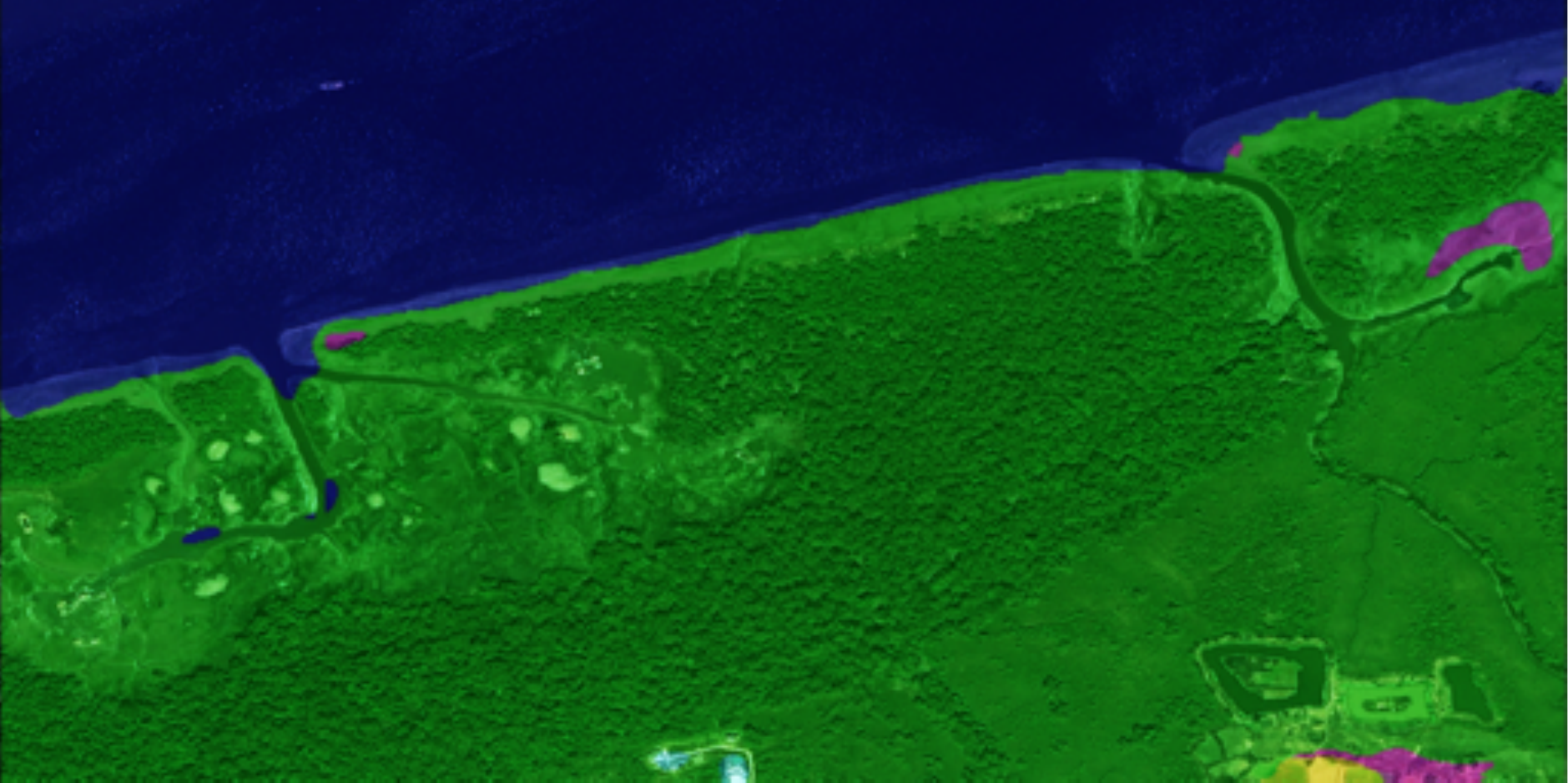}
    \end{subfigure}
    \begin{subfigure}[b]{\subfigwidth}
        \centering
        \makebox[\subfigwidth]{\small{(f) DenseCRF}}
    	\includegraphics[width=\columnwidth,height=\heightwidth]{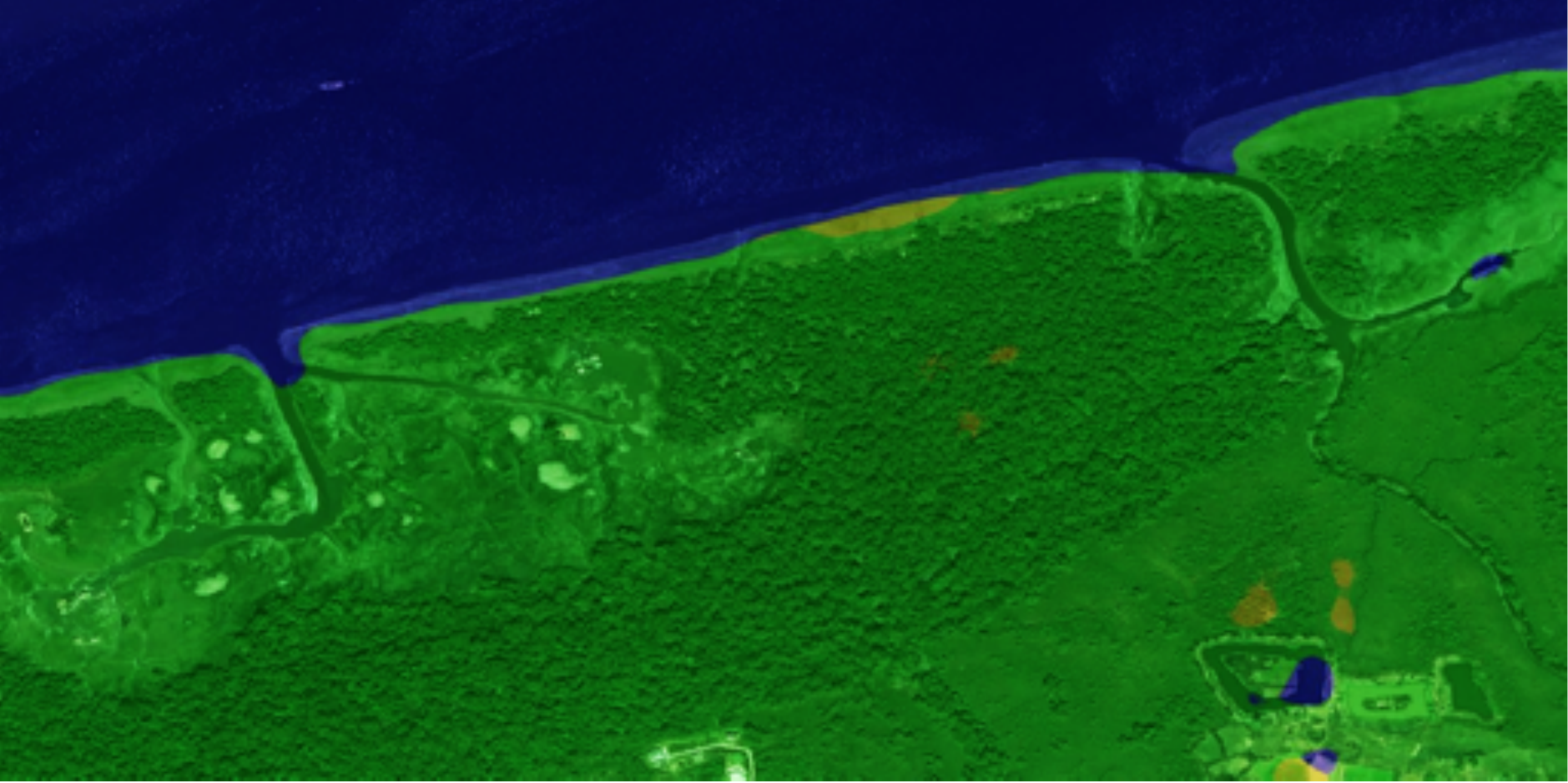}
    \end{subfigure}    
    \vspace*{2mm}
    
    \begin{subfigure}[b]{\subfigwidth}
        \centering
        \makebox[\subfigwidth]{\small{(g) PointRend}}
    	\includegraphics[width=\columnwidth,height=\heightwidth]{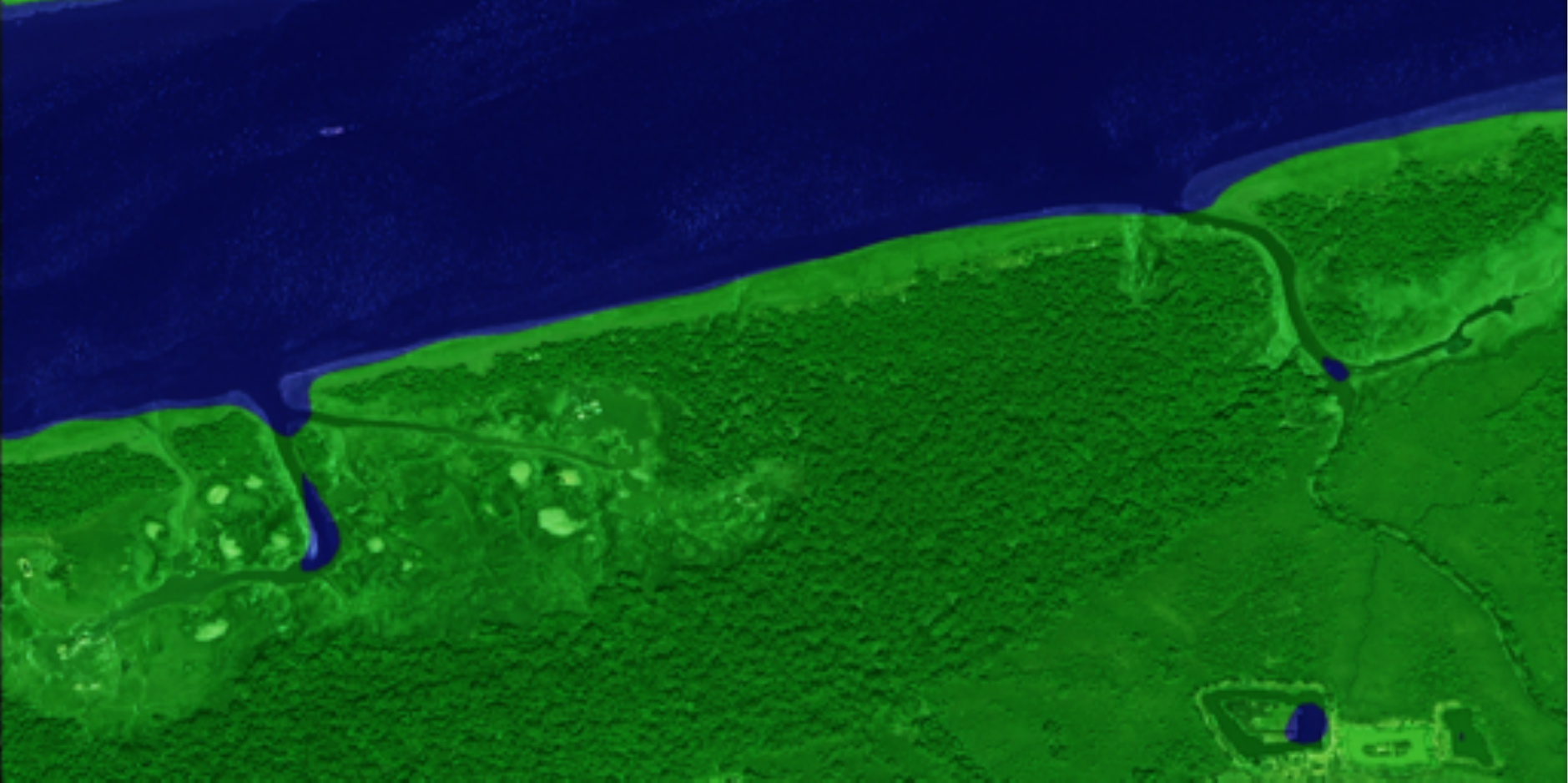}
    \end{subfigure}
    \begin{subfigure}[b]{\subfigwidth}
        \centering
        \makebox[\subfigwidth]{\small{(h) MagNet (Proposed)}}
    	\includegraphics[width=\columnwidth,height=\heightwidth]{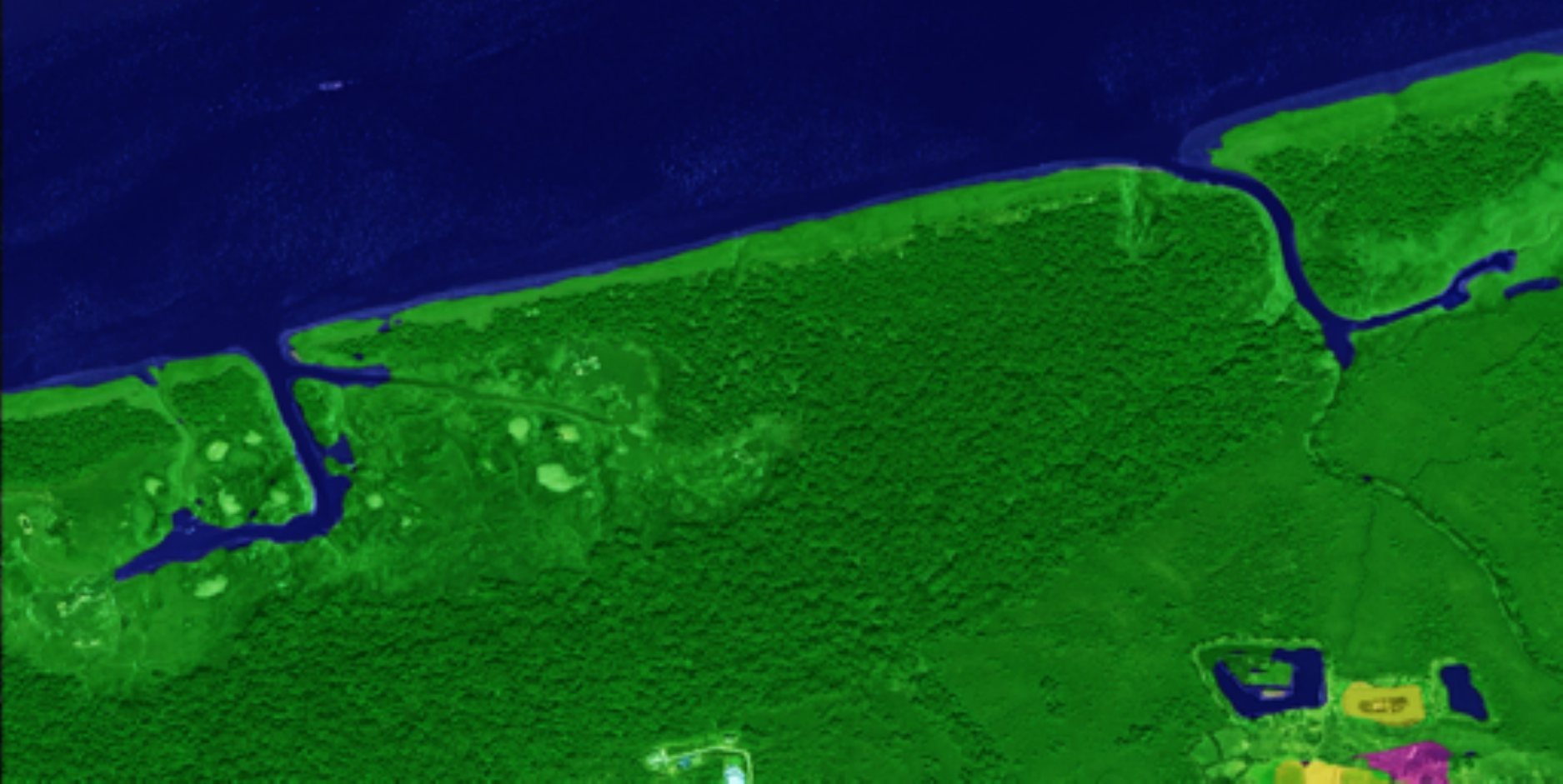}
    \end{subfigure}    
\vskip -0.05in
	\caption{\small {\bf Comparing several semantic segmentation and refinement approaches on a high-resolution input image}. Downsampling loses fine details, while Patch Processing wrongly classifies local patches due to the lack of the global context. The collaborative global-local network GLNet fails due to the large discrepancy between the global and local branches. Post-processing and refinement methods such as DenseCRF and PointRend  can only correct small mistakes due to local inconsistency. MagNet outperforms other methods, thanks to a novel multi-scale segmentation and refinement framework. 
	Best viewed in color.}
	\label{fig:teaser}
	\vspace{-2em}
\end{figure}

The current state-of-the-art (SOTA) semantic image segmentation techniques \cite{fcn,unet,segnet,chen2017deeplab,liu2019auto,m_Nguyen-etal-ICCV17,m_Le-etal-ECCV18} are based on deep learning, where a convolutional neural network (CNN) takes an input image and outputs a segmentation map. Most of the existing techniques, however, assume that the entire segmentation process can be performed with a single feed-forward pass of the input image and the entire process can be fitted into GPU memory. Unfortunately, most existing techniques cannot handle high-resolution input images due to memory and other computational constraints. 
One approach to handle a large input image is to downsample it, but this results in a low-resolution segmentation map, which is not adequate for applications that require high-resolution output with fine details~\cite{huynh2018context, nirkin2018face}, \textit{e.g.}, for tracking the progression of malignant lesion \cite{cuadros2009eyepacs}. Another approach to handle a large input image is to divide the image into small patches and process each patch independently. This approach, however, does not take into account the global information \cite{mottaghi2014role} that is needed to resolve ambiguity in local patches. The limitations of these two approaches are illustrated in \Fref{fig:teaser}(c) \& (d).

One way to address the limitations of the two aforementioned approaches is to combine them, i.e., to fuse global and local segmentation processes. On the one hand, the global view of the entire image can be used to resolve the ambiguity in the appearance of local patches. On the other hand, by analyzing local patches, we can refine the segmentation boundaries and recover the lost details due to the downsampling procedure of the global segmentation process. This approach has been successfully demonstrated recently by the Global-Local Network (GLNet)~\cite{chen2019collaborative}. However, given an input image with ultra-high resolution, there is a huge gap between the scale of the whole image and the scale of the local patches. This will lead to contrasting output segmentation maps, and it will be difficult to combine and reconcile differences with a single feed-forward processing stage (see~\Fref{fig:teaser}e); the difficulty of this combination task is analogous to constructing a single-span bridge across a wide river.

To bridge the gap between the two extreme ends of the scale space, we propose to consider multiple scales in between. We introduce a novel multi-scale framework where the output segmentation map will be progressively refined as the image is analyzed from the coarsest to the finest scale. 
The core of our framework is a refinement module that can use one segmentation map to refine another. This refinement module is used at every stage of our multi-scale processing pipeline to refine the output segmentation map at its most uncertain locations. Our framework can integrate global contextual cues to produce more accurate segmentation, and it can output high-resolution detailed segmentation maps under a memory constraint. \Fref{fig:teaser} shows the result of MagNet and compares it with other segmentation methods, including the recently proposed PointRend~\cite{pointrend} method that seeks to refine only at the most uncertain pixels.

\vspace{-0.25em}

\section{Related Work}
\vspace{-0.5em}
\begin{table*}[t]
    \vspace{-1em}
    \centering
    \begin{tabu}to \textwidth {@{}X[5.2,l] X[1.9,c] X[1,c] X[1,c] X[1,c] X[1.2,c] X[1.6,c] X[1.9,c] X[1.2,c] X[1.7,c] X[1,c]@{}}
    \toprule
         \multirow{2}{*}{Method} & 
         Dense & 
         GF & 
         DGF & 
         ISS & 
         GLNet &
         Cascade &
         PointRend &
         SegFix &
         DeepStrip &
         \multirow{2}{*}{Ours} \\
         & CRF \cite{densecrf} & 
         \cite{guidedfilter} & 
         \cite{deepguidedfilter} & 
         \cite{li2016iterative} & 
         \cite{chen2019collaborative} &
         PSP \cite{cascadepsp} &
         \cite{pointrend} &
         \cite{segfix} &
         \cite{deepstrip} \\
    \midrule
    Deep learning & 
    & 
    & 
    \cmark&
    \cmark&
    \cmark&
    \cmark&
    \cmark&
    \cmark&
    \cmark&
    \cmark\\
    
    Using high-resolution input &
    \cmark&
    \cmark&
    \cmark&
    &
    \cmark&
    \cmark&
    &
    \cmark&
    \cmark&
    \cmark\\
    
    Multi-scale processing &
    &
    &
    &
    &
    \cmark&
    \cmark&
    \cmark&
    &
    &
    \cmark\\
    
    Can recognize new objects &
    \cmark&
    \cmark&
    \cmark&
    \cmark&
    \cmark&
    \cmark&
    \cmark&
    &
    &
    \cmark\\
    
    Partly refinement &
    &
    &
    &
    &
    &
    &
    \cmark &
    \cmark &
    \cmark &
    \cmark\\
    \bottomrule
    \end{tabu}
    \caption{Summary of key features of various semantic segmentation refinement approaches.}
    \label{tab:method_sum}
    \vspace{-1em}
\end{table*}



\myheading{Multi-scale, multi-stage, context aggregation.} The combination of multiple scale levels helps the network aggregate different fields of view and provides more context to each pixel \cite{chen2016attention,m_Hou-etal-PAMI21}. ICNet~\cite{zhao2018icnet} used a cascaded architecture for feature maps of different downsampled inputs, while RefineNet~\cite{lin2017refinenet} fused upsampled outputs of the branches that handled different low-resolution inputs. Feature Pyramid Network (FPN)~\cite{fpn} upsampled feature maps in different scales and aggregated them with the output of low layers. The dilated convolution and Atrous Spatial Pyramid Pooling (ASPP) module in DeepLab~\cite{chen2017deeplab} enlarged the receptive field, created a connection between far-apart pixels. The same effect was achieved by PSPNet~\cite{zhao2017pyramid}, which combined different scaling feature maps to enlarge the receptive fields. High-resolution Net (HRNet)~\cite{wang2020hrnet} proposed another scale fusion schema where a new branch with a larger receptive field was added after each stage. Attention technique was also used in many recent approaches \cite{ocr, contextprior2020, blendmask2020} to add more global information to every single point.

Another approach to handle high-resolution images is to use multi-stage networks, where images are segmented after several stages or sub-networks. \citet{xia2016zoom} proposed Hierarchical Auto-Zoom Net, a strategy to scale the field of view when sliding the view through a big image. For ultra-high resolution images, \citet{takahama2019multi} solved the imbalance between background and foreground by predicting whether the whole patch contained foreground pixels or not before segmenting it. 

Propagating global information to local patches is a promising approach to deal with high-resolution images. ParseNet~\cite{liu2015parsenet} pooled global context to local field of view to have more information. BiSeNet~\cite{yu2018bisenet} included one more branch for global pooling and the global context would be added to the feature map at the last stage. Although those methods were efficient, they consumed a huge amount of GPU memory for ultra-high resolution images. \citet{tokunaga2019adaptive} proposed a method for super-high-resolution images, using independent multi-scale networks and an adaptive weight generator. The outputs of network members were combined with corresponding trained weights to produce the final output, but there was no knowledge sharing between network branches. Unlike \cite{tokunaga2019adaptive}, \cite{chen2019collaborative} contained two sub-networks with shared information, where the global branch took the downsampled images to extract global context and the local branch took patches and corresponding global features to improve the details of high-resolution images. However, due to the ad-hoc combination of the global and local branches, it was difficult to extend to more than two scales.  Moreover, in our experience, training the local network was difficult due to the domination of the strong global branch. 

\myheading{Segmentation refinement.} There were several approaches to improve the segmentation outputs with post-processing. One approach was to use classical methods such as Conditional Random Fields (CRFs)~\cite{densecrf} or Guided Filter (GF)~\cite{guidedfilter} on the segmentation mask produced by deep learning networks. However, these methods were slow and the improvement was incremental. The inference speed could be improved with a deep learning version of Guided Filter (DGF)~\cite{deepguidedfilter}. Another approach for post-processing was to use deep networks. Iterative Instance Segmentation (ISS)~\cite{li2016iterative} refined the output by repeatedly passing the input image and the segmentation map through a refinement module several times. This method was based on self-reflection, the input image to each refinement stage was the same. Like ISS, CascadedPSP~\cite{cascadepsp} used the same refinement scheme but the resolution of the input at each refinement stage was different. However, the wrong prediction at any middle stage could significantly affect the performance of later steps. Some methods aimed to refine parts of the output only, such as pixels at the boundaries~\cite{deepstrip,segfix} or pixels at uncertain locations (PointRend~\cite{pointrend}). However, boundary refinement methods~\cite{deepstrip,segfix} failed to recover tiny objects, while PointRend~\cite{pointrend} only used the local context for refinement. 
Furthermore, because the input of the PointRend was the high-level features of a deep network, it must be trained specifically for each segmentation backbone. In this paper, we propose a modular framework for having any number of scale levels. It is simple but effective for refining a coarse segmentation output, being able to keep the overall structure of the coarse segmentation output while adding more details after each stage without suffering the domination problems. \Tref{tab:method_sum} compares the key features of different methods.

\vspace{-0.25em}

\section{MagNet}
\vspace{-0.25em}
\begin{figure*}[h]
    \vspace{-3.0em}
	\centering \includegraphics[width=\textwidth]{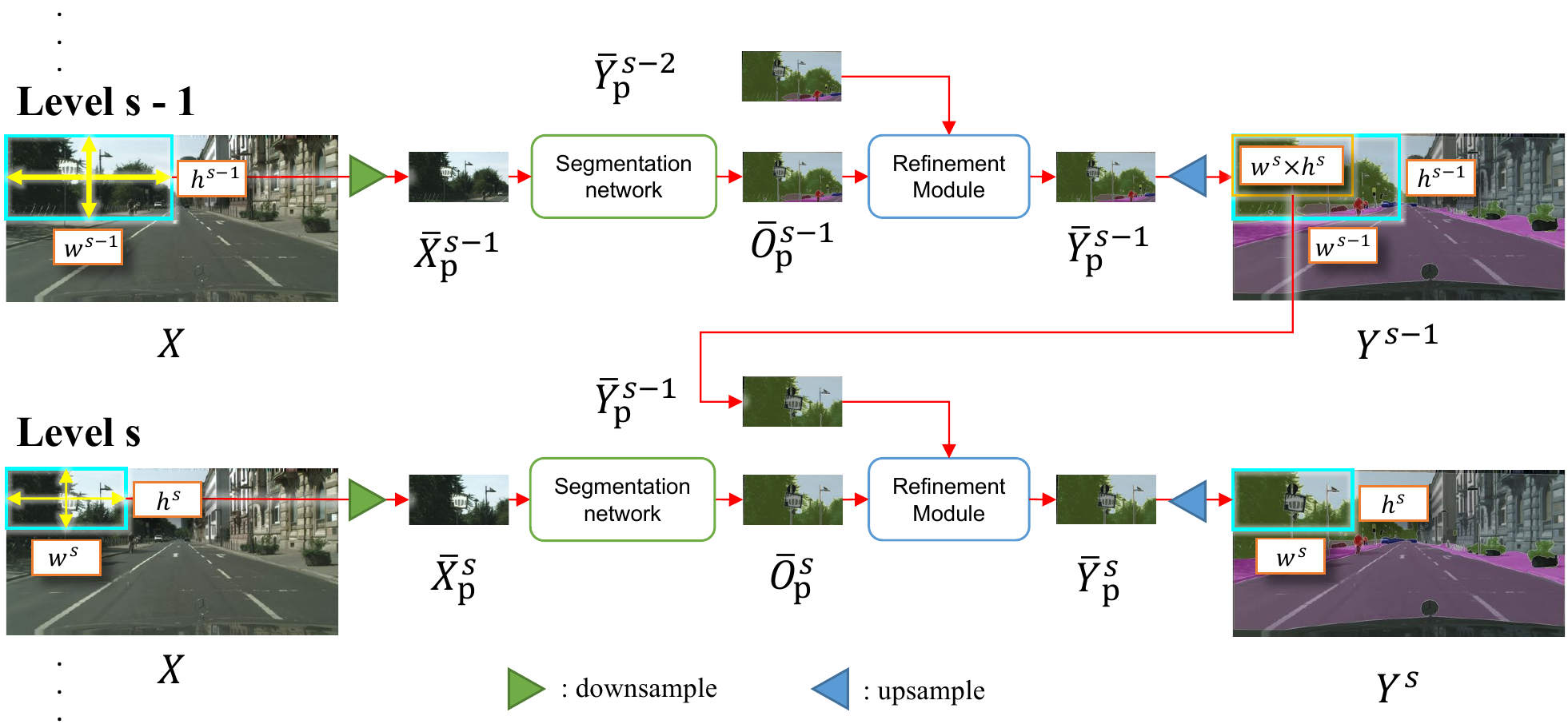}
	\caption{{\bf Overview of our proposed MagNet}. The segmentation network produces the scale-specific prediction while the refinement module selectively refines the coarse prediction from previous stages based on that local prediction.}
	\label{fig:overall_structure}
	\vspace{-1em}
\end{figure*}

We now describe MagNet, a multi-scale segmentation framework for ultra-high resolution images. It is a multi-stage network architecture, where each stage of the network corresponds to a particular scale. An input image will be inspected at multiple scales, from the coarsest to the finest. 

The core of our framework is a segmentation module and a refinement module, which are used at every processing stage. At each stage, the inputs to the refinement module are two segmentation maps: (1) the cumulative result from the previous stages, and (2) the result obtained by running the segmentation module at and only at the current scale. The objective of the refinement module is to use the latter segmentation map to refine the former one, at selective locations determined based on the uncertainty of two estimated segmentation maps. 

In our framework, the segmentation module can be any segmentation backbone, as long as it can output a segmentation map with uncertainty estimates. The refinement module is agnostic to the segmentation backbone; it can be trained with one backbone and used with another. In the following, we will describe the multi-stage processing pipeline and the refinement module in details. 

\subsection{Multi-stage processing pipeline}
\label{sec:multinetwork}

The architecture and processing pipeline of MagNet is depicted in \Fref{fig:overall_structure}. There is one segmentation module and one refinement module, which are used repeatedly in $m$ processing stages, where $m$ is a hyper-parameter for the number of scales that we want to analyze. We use $s$ to denote the processing stage, where $s=1$ corresponds to the coarsest scale and $s=m$ corresponds to the finest scale. Let $X \in \mathbb{R}^{H{\times}W{\times}3}$ be an input image, where $H, W$ are the height and width of the image. We consider the case when $H$ and $W$ are too big for image $X$ to be processed without downsampling, and let $h$ and $w$ be the largest  (or desirable) sizes that can be handled by the segmentation module. We use $h^s$ and $w^s$ to denote the height and width for the scale level $s$. We determine the scale levels so that they span the entire scale space: $H = h^1 > \cdots > h^m = h$ and $W = w^1 > \cdots > w^m = w$.

For a particular scale level $s$, we divide the input image $X$ into patches of size $\htimesw{h^s}{w^s}$ and perform semantic segmentation on these patches. The locations of these patches are defined by a set of rectangular windows, and let $\mP^s$ denote the set of these windows: $\mP^s = \{\p| \p = (x, y, w^s, h^s) \}$, where each window is specified by the top-left corner, width, and height. As the scale level $s$ increases, the width and height of the rectangular windows decrease, but the cardinality of $\mP^s$ increases. For a particular window $\p$, we will use $X_{\p}$ to denote the image patch extracted at the window $\p$. 

Our network will take an image $X \in \mathbb{R}^{H{\times}W{\times}3}$ and produce a sequence of segmentation maps $Y^1, \cdots, Y^m \in \mathbb{R}^{H{\times}W{\times}C}$, where $C$ is the number of semantic categories in consideration. At stage $s$, we first determine the set of rectangular windows  $\mP^s$ for patch division and refine the segmentation map of each patch. Specifically, for each window $\p \in \mP^s$, do:
\begin{enumerate} \denselist 
        \item Extract the image patch $X^s_{\p}$ and previous segmentation output $Y^{s-1}_\p$ defined by the window $\p$. The height and width of these tensors are $h^s$ and $w^s$.
        \item Downsample $X^s_{\p}$ and $Y^{s-1}_\p$ so that the new height and width are $h$ and $w$, which are the size that can be fitted into GPU memory and be processed by the segmentation and refinement modules. Let $\bar{X}^s_{\p}$ and $\bar{Y}^{s-1}_{\p}$ denote the downsampled tensors.
        \item With $\bar{X}^s_{\p}$ as the input, use the segmentation module to obtain the scale-specific segmentation map $\bar{O}^s_\p$.
        \item With $\bar{Y}^{s-1}_{\p}$ and $\bar{O}^s_\p$ as the input, use the refinement module to refine $\bar{Y}^{s-1}_{\p}$ to obtain $\bar{Y}^s_{\p}$ (See \Sref{sec:refinement_module}). 
        \item Upsample $\bar{Y}^s_{\p}$ to get $Y^s_{\p}$ of size $h^s{\times}w^s{\times}C$.
\end{enumerate}
These processing steps are illustrated in \Fref{fig:overall_structure}. 

\subsection{Refinement module \label{sec:refinement_module}}
The refinement module is a core component of our framework, which is used to refine the individual patches of a segmentation map at every processing stage of our pipeline. The input to this module is two segmentation maps of size $h{\times}w{\times}C$: (1) the scale-cumulative segmentation map $Y$, from all previous scales, and (2) the scale-specific segmentation map $O$, from the current scale. The output of the module is the updated scale-cumulative segmentation map. 

\begin{figure}[!t]
    \vspace{-1.0em}
    \centering
    \includegraphics[width=\columnwidth]{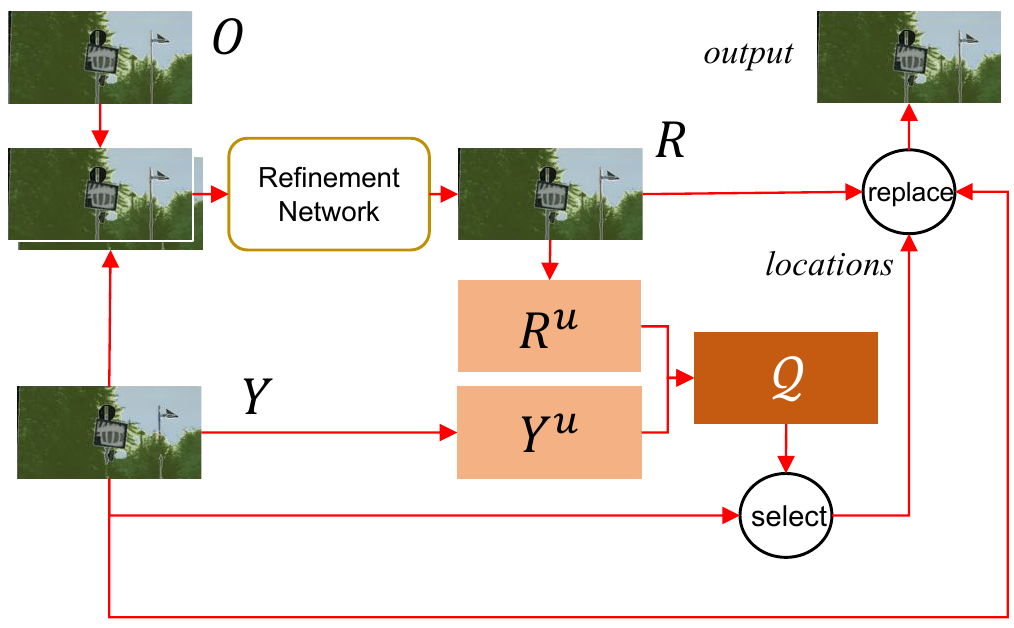}
    \caption{{\bf The overview of the refinement module}. The cumulative segmentation $Y$ is partly replaced with the scale-specific segmentation map $O$ based on the score $\mQ$.}
    \label{fig:refinement_module}
    \vspace{-1.5em}
\end{figure}

\Fref{fig:refinement_module} depicts the refinement process, which contains the following steps. First, using a small network with $Y$ and $O$ as the input, we obtain an initial combined segmentation map $R$. We then calculate the prediction uncertainty maps for both $Y$ and $R$.  Specifically, for each pixel of $Y$, the prediction confidence at this location is defined as the absolute difference between the highest probability value and the second-highest value (among the $C$ probability values for $C$ classes). The uncertainty score is then computed based on the confidence score such that the two scores must add up to one. Similarly, we can compute the prediction uncertainty map for $R$. Let $Y^u$ and $R^u$ denote the prediction uncertainty maps for $Y$ and $R$ respectively. 

Next, we will describe how to use two uncertainty maps to select $k$ locations of $Y$ for refinement. While only one uncertainty prediction is used in the previous work PointRend~\cite{pointrend}, we extend this approach to have a better selection strategy. These are the locations where $Y$ is uncertain about its prediction, while $R$ is certain about its prediction. The score map for ranking the pixels is calculated as $\mQ = \mF(Y^u \odot (1 - R^u))$, where $\odot$ denotes point-wise multiplication and $\mF$ denotes median blurring to smooth the score map. Empirical comparison between the effect of each element in the formula can be found in \Sref{sec:ablation_selection}.

\vspace{-0.5em}
\subsection{MagNet-Fast}

There is trade-off between the accuracy and the run-time efficiency of our framework. One way to reduce the running time is to decrease the number of scales to process. Another approach is to perform segmentation and refinement on a subset of image patches at each scale level. MagNet-Fast combines these two approaches when it runs on a smaller number of scales and only selects the patches with the highest prediction uncertainty $Y^u$ for refinement. In MagNet-Fast, the total number of image patches that need to be fed into the segmentation module might be even smaller than the number of image patches used in the patch processing approach. Moreover, MagNet-Fast can leverage both global context and detailed information for segmentation, leading to superior results as will be seen in our experiments. 
\vspace{-0.25em}

\section{Experiments}
\vspace{-0.25em}
We evaluated the performance of MagNet on three high resolution datasets: Cityscapes~\cite{cityscapes}, DeepGlobe~\cite{demir2018deepglobe}, and Gleason~\cite{gleason2019}. Some information about these datasets is listed in \Tref{tab:datasets}. The number of pixels of each image is from 2 to 25 million. In this section, we present experiments comparing MagNet with other state-of-the-art frameworks in semantic segmentation and also describe some ablation studies on Cityscapes.

\begin{table}[h!]
    \centering
    \begin{tabu} to \columnwidth {@{}X[3,l]X[3,l]X[2,c]X[2,c]@{}}
    \toprule
    Dataset & Content & Resolution & No.classes \\
    \midrule
    Cityscapes \cite{cityscapes} & urban scene & $2048{\times}1024$ & 19 \\ 
    DeepGlobe \cite{demir2018deepglobe} & aerial scene & $2448{\times}2448$ & 6 \\
    Gleason \cite{gleason2019} & histopathology & $5000{\times}5000$ & 4 \\
    \bottomrule
    \end{tabu}
    \vskip -0.05in
    \caption{{\bf Details of high-resolution datasets used to evaluate our framework.} All images have from 2 to 25 million pixels with a lot of details.}
    \label{tab:datasets}
    \vspace{-1em}
\end{table}

\subsection{Implementation details}

\begin{figure}
    \vspace{-1em}
    \centering
    \includegraphics[width=\columnwidth]{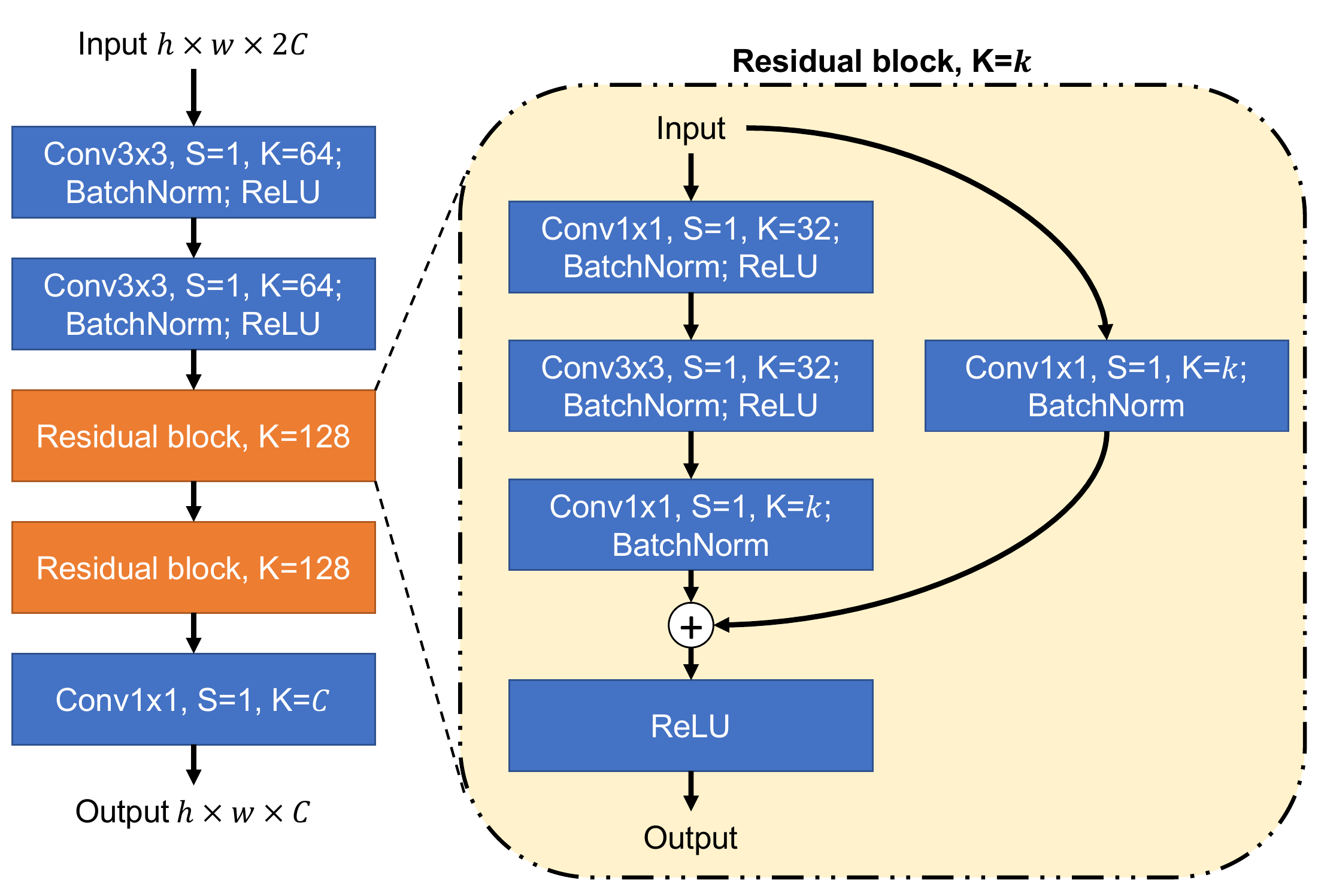}
    \vskip -0.1in
    \caption{{\bf The two residual blocks are trained to refine the segmentation at each scale}. This module outputs the same size $h{\times}w$ as the input.}
    \label{fig:refinement_network}
    \vspace{-1.5em}
\end{figure}

\vspace{-0.5em}
\myheading{Architecture of the refinement module.} \Fref{fig:refinement_network} depicts the architecture of the refinement module used in all experiments. The main components are the two residual blocks. With the input of size $h{\times}w{\times}2C$, the refinement module produces the output of size $h{\times}w{\times}C$.

\myheading{Training.} For each dataset, we trained a state-of-the-art segmentation model on the downsampled images and a refinement module to refine the coarse output on sliced images. While training the refinement module, we randomly extracted image patches and also applied the following data augmentation: rotation, and horizontal and vertical flipping. We used SGD optimizer with momentum $0.9$, decayed weight $5{\times}10^{-4}$, and initial learning rate $10^{-3}$. We trained the refinement module for 50 epochs, and the learning rate was decreased tenfold at epoch 20, 30, 40, and 45. We used cross-entropy as the loss function for training segmentation and refinement modules. We implemented MagNet using PyTorch \cite{paszke2019pytorch} starting from the public implementation of HRNet-OCR~\cite{ocr}. We use a batch size of 16 for training on a DGX-1 workstation with Tesla V100 GPUs.

\myheading{Testing.} During inference, at each scale, we extracted non-overlapping patches for processing. The evaluation metric is mean Intersection over Union (mIoU). For memory and speed comparison, we ran benchmarking experiments on a machine with an Intel i7 CPU and an RTX2080Ti GPU.

\subsection{Experiments on the Cityscapes dataset}
Cityscapes is a dataset of high-resolution urban scenes, containing images of size $2048{\times}1024$ pixels. The task is to segment objects in videos captured by auto cameras. There are two kinds of data, with coarse and fine labels. In our experiments, we used the fine-labeled dataset, with 2,975 training images and 500 testing images.


We considered four possible scale levels ($m{=}4$), corresponding to patch sizes of $2048{\times}1024$, $1024{\times}512$, $512{\times}256$, and $256{\times}128$. The size of the input to the segmentation module was always $256{\times}128$ to satisfy the memory constraint, so any larger patch would be downsampled. 
\vspace{-1em}

\subsubsection{Benefits of multiple scale levels}
\vspace{-0.5em}

\begin{table}[t]
    \centering
    \begin{tabu} to \columnwidth {@{}X[3,l]X[1,c]X[1,c]@{}}
    \toprule
    Refinement steps & mIoU(\%) & Time(s) \\
    \toprule
    256  & 63.23 & 0.03 \\
    256$\rightarrow$512 & 65.73 & 0.19 \\
    256\hspace{2.75ex}$\rightarrow$\hspace{2.75ex}1024 & 65.23 & 0.61 \\
    256\hspace{6ex}$\rightarrow$\hspace{6ex}2048 & 65.21 & 2.24\\
    256$\rightarrow$512\hspace{3.25ex}$\rightarrow$\hspace{3.25ex}2048 & 67.13 & 2.38 \\
    256\hspace{2.75ex}$\rightarrow$\hspace{2.75ex}1024$\rightarrow$2048 & 66.95 & 2.79 \\
    256$\rightarrow$512$\rightarrow$1024$\rightarrow$2048 & \textbf{67.57} & 2.93 \\
    \bottomrule
    \end{tabu}
    \vskip -0.1in
    \caption{{\bf Performance of MagNet on Cityscapes with and without intermediate scale levels}. It is essential to have the intermediate scales.}
    \label{tab:cityscapes_scales}
    \vspace{-1.5em}
\end{table}

\begin{figure}[!b]
    \centering
    \vspace{-0.5em}
    \includegraphics[width=\columnwidth]{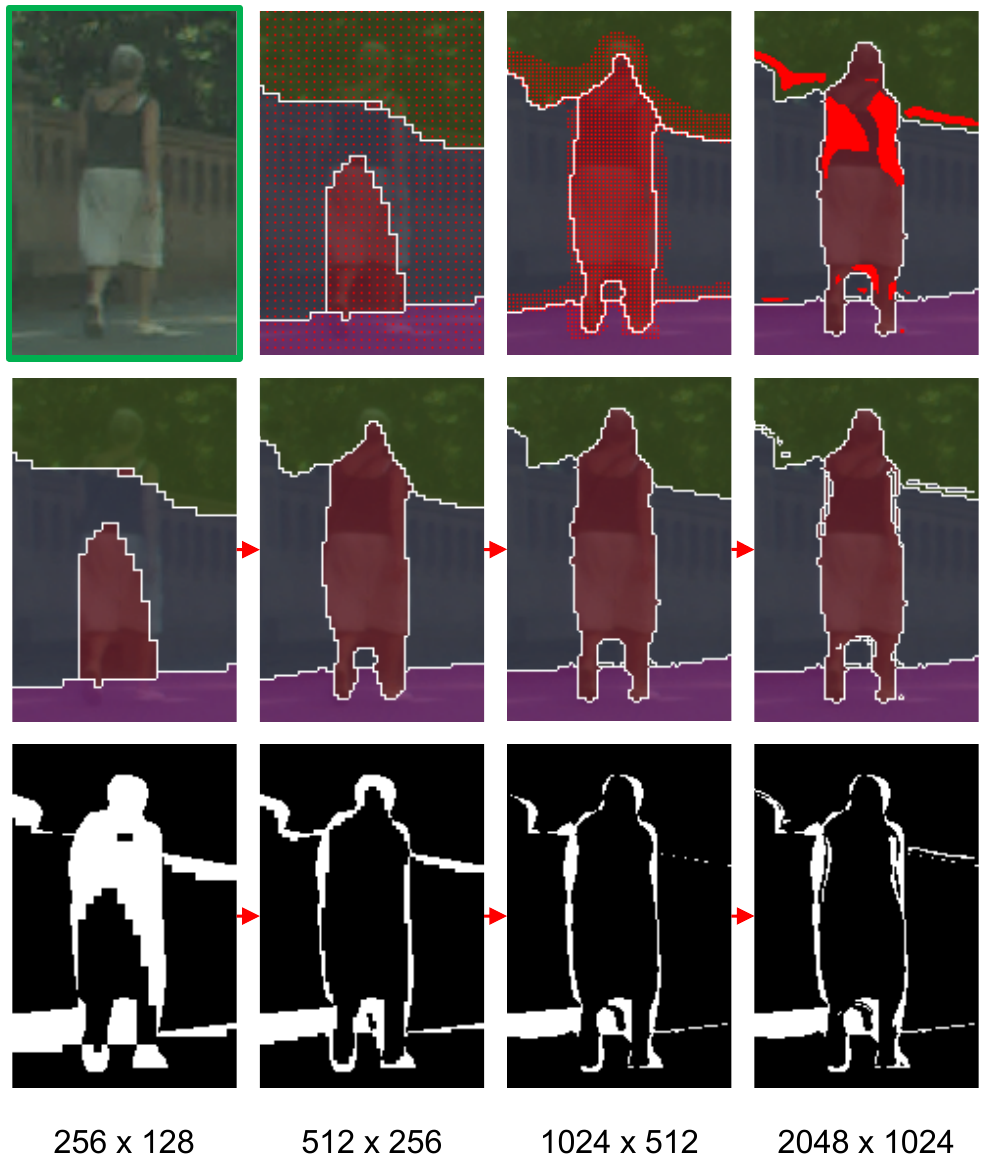}
    \vskip -0.1in
    \caption{\textbf{The visualization of segmentation output through each processing stage on Cityscapes dataset}. The first one is the image. Others in the first row are the selective points to be refined (red color). The second row is the segmentation output. The third row is the errors comparing with the ground-truth. Best viewed in color. }
    \label{fig:step_refinement}
    \vspace{-0.75em}
\end{figure}

\Tref{tab:cityscapes_scales} shows the results of MagNet for a different number of scales.  While the direct refinement from the lowest to highest scale improves about 2\%  mIoU, from 63.23\% to 65.23\%, adding the two intermediate scales between the smallest and largest scales improve the performance by 4.34\% mIoU. Qualitative improvements through each processing stage can be observed in \Fref{fig:step_refinement}. After each stage, the errors decrease and the segmentation masks become finer.
\vspace{-0.25em}
\subsubsection{Comparing segmentation approaches \label{sec:com_approach}}
\vspace{-0.5em}

\begin{figure*}[t!]
    \vspace{-0.75em}
    \centering
    \begin{subfigure}[b]{0.32\textwidth}
        \centering
        \includegraphics[width=\textwidth]{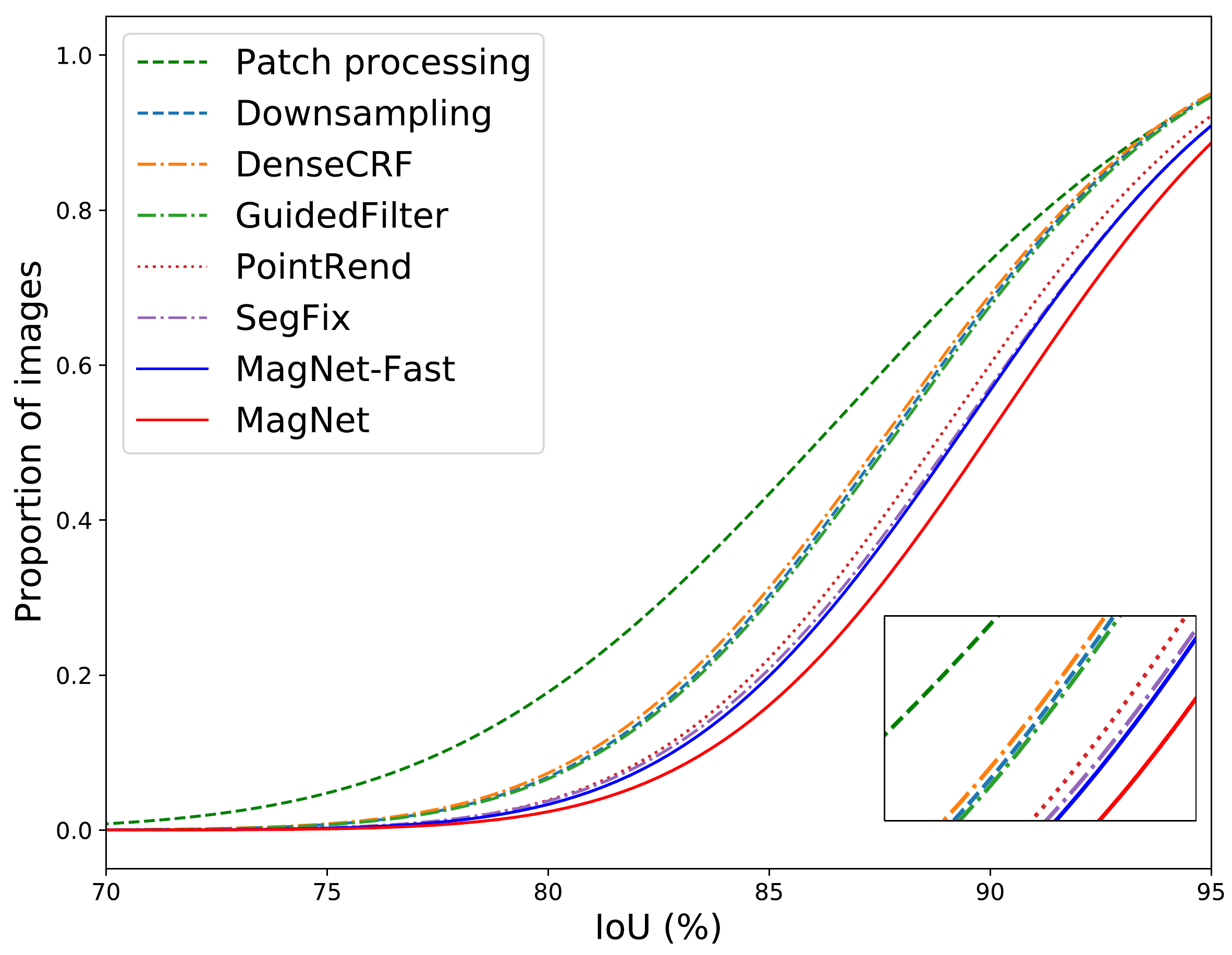}
        \caption{The IoU cumulative distribution}
        \label{fig:cityscapes_result:dist}
    \end{subfigure}
    \begin{subfigure}[b]{0.66\textwidth}
        \centering
        \includegraphics[width=\textwidth]{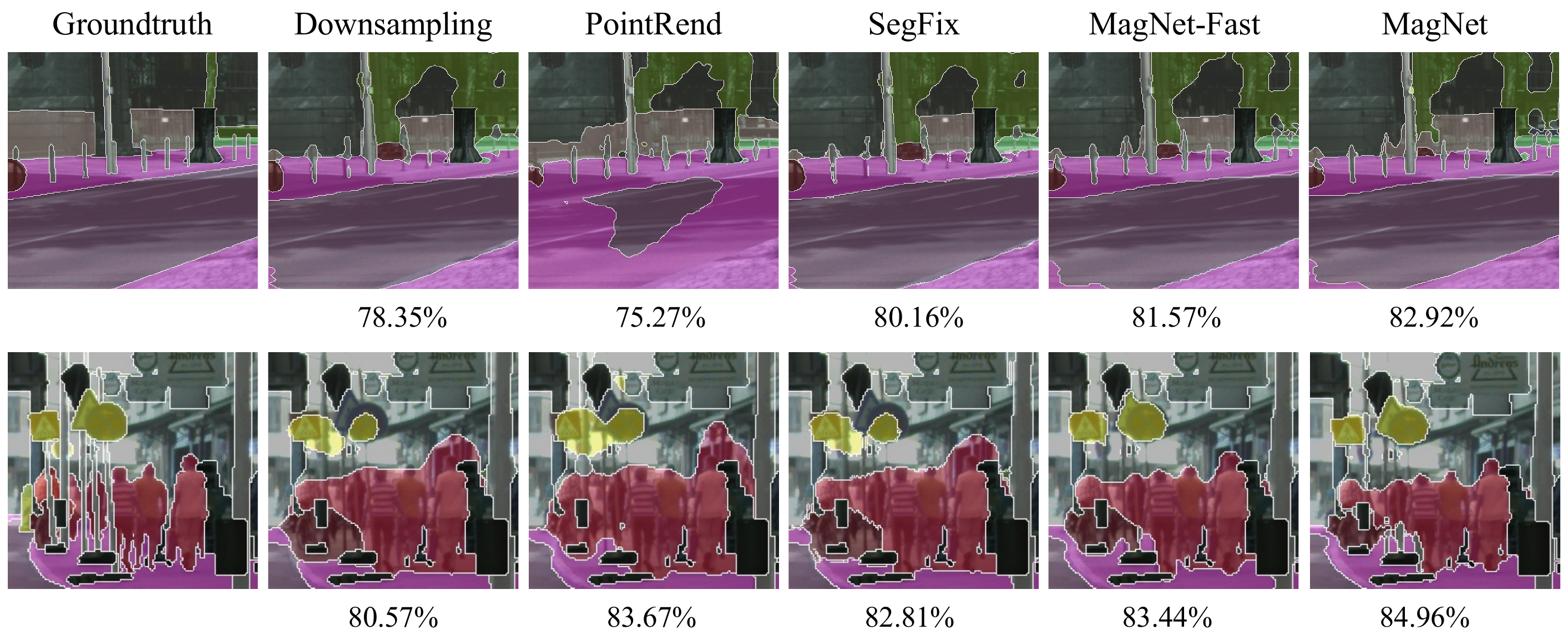}
        \caption{Visualization of refinement methods}
        \label{fig:cityscapes_result:vis}
    \end{subfigure}
    \vskip -0.05in 
    \caption{{\small \textbf{Our methods outperform other refinement frameworks on the Cityscapes dataset}. (a) The cumulative distribution of mIoU of each image on the dataset (lower is better). The MagNet and MagNet-Fast achieve the best result among others. (b) Some segmentation results of refinement frameworks and our MagNet. The mIoU numbers are below the images. More tiny objects are recognized and boundaries are refined better with MagNet and MagNet-Fast. Best viewed in a digital device with magnification.}}
    \label{fig:cityscapes_result}
    \vspace{-1.0em}
\end{figure*}

\newcommand{\smallup}[1]{\scriptsize{($\uparrow${#1})}}
\newcommand{\smalldown}[1]{\scriptsize{($\downarrow${#1})}}
\begin{table}[!t]
    \centering
    \begin{tabu} to \columnwidth {@{}X[3.5,l]X[3,l]X[2.5,l]X[1.5,r]@{}}
    \toprule
    Model & mIoU(\%) & Mem.(MB) & Time(s) \\
    \midrule
    Patching & 52.19 & 1575 & 1.77 \\
    Downsampling  & 63.23 & 1575 & 0.02 \\
    DenseCRF~\cite{densecrf} & 62.95~\smalldown{0.28} & 1575 & 26.02 \\
    DGF~\cite{deepguidedfilter} & 63.33~\smallup{0.10} & \textbf{1727}~\smallup{152} & 0.32 \\
    PointRend~\cite{pointrend} & 64.39~\smallup{1.16} & 2033~\smallup{458} & \textbf{0.14} \\
    SegFix~\cite{segfix} & 65.83~\smallup{2.60} & 2961~\smallup{1386} & 0.38\\
    MagNet-Fast & 66.91~\smallup{3.68} & 2007~\smallup{432} & 0.32 \\
    MagNet & \textbf{67.57}~\smallup{4.34} & 2007~\smallup{432} & 2.93\\
    \bottomrule
    \end{tabu}
    \vskip -0.1in
    \caption{{\bf Performance of MagNet and other segmentation refinement methods on the Cityscapes dataset}. The backbone HRNetV2-W18+OCR \cite{ocr} was used as the segmentation module for all refinement methods. }
    \label{tab:cityscapes_result}
    \vspace{-0.75em}
\end{table}

\newcommand{\best}[1]{\textcolor{red}{\textbf{#1}}}
\newcommand{\second}[1]{\textcolor{blue}{\textbf{#1}}}
\begin{table}[t]
    \centering
    \vspace{-0.25em}
    \begin{tabu} to \columnwidth {@{}X[2.0,l]X[1,c]X[1,c]X[1,c]X[1,c]X[1,c]@{}}
    \toprule
    Model
    & fence & pole & traffic sign & person & motor-cycle \\
    \midrule
    Patching & 33.32 & \second{42.87} & 59.39 & 61.23 & 22.12\\
    Downsample & 45.38 & 35.58 & 54.97 & 64.64 & 36.16 \\
    DenseCRF~\cite{densecrf} & 45.30 & 32.27 & 54.69 & 64.32 & 36.20 \\
    DGF~\cite{deepguidedfilter} & 45.42 & 35.63 & 55.27 & 64.85 & 36.20 \\
    PointRend~\cite{pointrend} & 45.01 & 42.71 & \second{60.18} & 67.10 & \second{39.17} \\
    SegFix~\cite{segfix} & \second{46.17} & 38.77 & 59.23 & \second{67.86} & 37.12 \\
    \midrule
    \small{MagNet-Fast} & 48.57 & 46.38 & \best{64.84} & 69.65 & 41.98 \\
    MagNet & \best{50.59} & \best{49.39} & 64.15 & \best{72.16} & \best{45.19}\\
    \midrule
    Improvement & 4.42 & 6.52 & 4.66 & 4.30 & 6.02\\
    \bottomrule
    \end{tabu}
    \vskip -0.1in 
    \caption{{\bf IoU for some specific categories on Cityscapes.} The best and previous best method is highlighted in red and blue color respectively, and the difference between them is shown in the last row.}
    \label{tab:cityscapes_details}
    \vspace{-1.6em}
\end{table}

\Tref{tab:cityscapes_result} and \Tref{tab:cityscapes_details} compare the performance of MagNet with several state-of-the-art semantic segmentation refinement methods. All methods were trained and applied to the output of the pretrained HRNet+OCR~\cite{ocr} model, which is among the leading methods on this dataset. Although there are various HRNet models \cite{wang2020hrnet}, we used HRNetV2-W18 given its manageable complexity that was required for our experiments, especially in terms of memory constraint. 

MagNet-Fast is an efficient version of MagNet, in which only the most uncertain patches are refined at each scale. In this experiment, we selected the number of patches so that MagNet-Fast had a similar processing speed as SegFix~\cite{segfix} and DGF~\cite{deepguidedfilter}. This model ran on only three scales $256{\rightarrow}512{\rightarrow}2048$ and the three highest uncertain patches for each scale. In total, MagNet-Fast needed to run inference on $1 + 3 + 3 = 7$ patches, comparing to $1 + 4 + 16 + 64 = 85$ patches of MagNet, and 64 patches of the patch processing approach.

In this experiment, except for DenseCRF~\cite{densecrf}, we finetuned other frameworks to achieve the best result with the segmentation backbone. For SegFix~\cite{segfix}, with the offset prediction published by the authors, the best result was achieved with the offset width of 10. Both DGF~\cite{deepguidedfilter} and PointRend~\cite{pointrend} were trained with the output of the segmentation backbone. Besides, DGF ran on patches to be fairly compared with our method in speed and memory.

As can be observed, DenseCRF~\cite{densecrf} is the only method that cannot improve the coarse segmentation. PointRend~\cite{pointrend} is the fastest method, but the improvement is small. The inference times of MagNet-Fast, SegFix~\cite{segfix}, and DGF~\cite{deepguidedfilter} are similar, but MagNet-Fast outperforms the others significantly. MagNet was slower, but it had the highest mIoU.   

The cumulative IoU distributions of these methods in our experiment are shown on \Fref{fig:cityscapes_result:dist}. There is a big gap between MagNet and the other methods, especially when looking at the zoom-in window. 

\Fref{fig:cityscapes_result:vis} shows the results of several methods on two Cityscapes images. Both MagNet and MagNet-Fast yield the best refinement. SegFix~\cite{segfix} cannot recover small objects, such as sign poles, that are wrongly merged with a bigger region, while PointRend~\cite{pointrend} performs poorly due to the lack of global context. 
\vspace{-1em}

\subsubsection{Ablation study: point selection}
\label{sec:ablation_selection}
\vspace{-0.5em}

\begin{table}[t!]
    \centering
    \begin{tabu} to \columnwidth {@{}X[1,c]X[1,c]X[1,c]X[1,c]@{}}
    \toprule
    $Y^u$ & $(1 - R^u)$ & $\mF$ & mIoU (\%) \\
    \toprule
    \cmark &        &        & 63.22 \\
    \cmark &        & \cmark & 63.25 \\
           & \cmark &        & 66.36 \\
           & \cmark & \cmark & 66.46 \\
    \cmark & \cmark &        & 67.37 \\
    \cmark & \cmark & \cmark & \textbf{67.57} \\
    \bottomrule
    \end{tabu}
    \vskip -0.1in
    \caption{{\small {\bf Performance of MagNet on Cityscapes with different ranking scores}. The initial segmentation has mIoU of $63.23\%$. With $k{=}2^{16}$, the framework achieves the best performance when using both $Y^u$ and $(1 - R^u)$ with smoothing operation.}}
    \label{tab:cityscapes_ranking_score}
    \vspace{-0.75em}
\end{table}

\begin{figure}[h!]
    \vspace{-0.5em}
    \centering
    \includegraphics[width=\columnwidth]{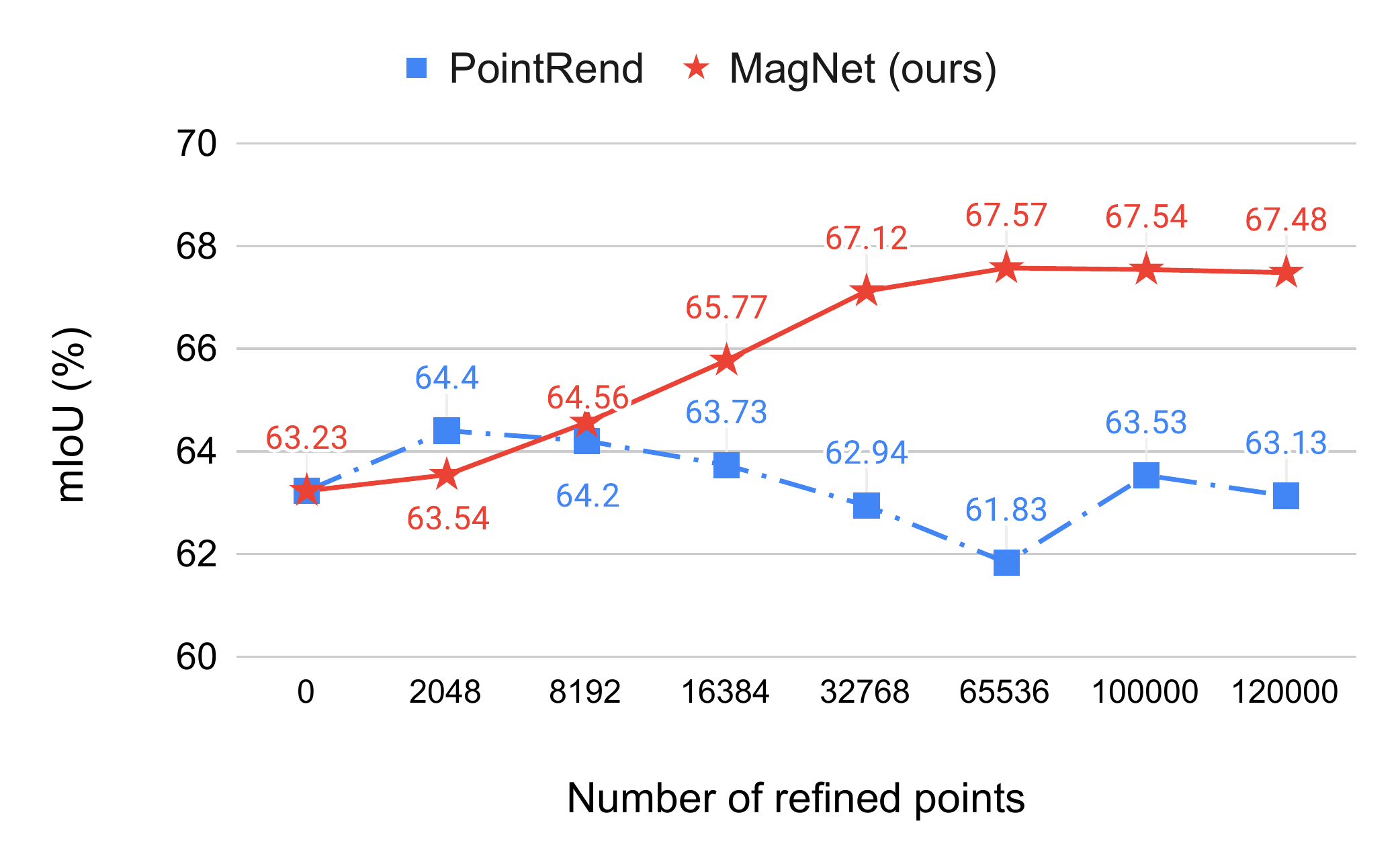}
    \vskip -0.1in
    \caption{ \textbf{Correlation between the number of selective points of each scale and mean IoU on the Cityscapes dataset}. When the quantity of points increases, the performance of MagNet continuously grows while the mIoU of PointRend decreases.}
    \label{fig:point_comparison}
    \vspace{-1.0em}
\end{figure}

\Tref{tab:cityscapes_ranking_score} shows our ablation study on the importance of using the prediction uncertainty maps $Y^u$, $R^u$, and the median filtering function $\mF$ in selecting points for refinement. The best performance was achieved when both uncertainty maps were used. Also, smoothing with median filtering improved the result in every case.

We also studied how the number of refinement points correlates with accuracy. The results of MagNet and PointRend~\cite{pointrend} are shown in \Fref{fig:point_comparison}. As can be seen, the performance of MagNet improved when the number of points increased. The performance stopped increasing after $2^{16}$ points, and it dropped to 66.86\% if all points were selected for refinement. Meanwhile, the performance of PointRend decreased significantly when the number of selected points increased beyond 2048; it even dipped below the initial value where no refinement was applied. 
\vspace{-1em}

\subsubsection{Ablation study: segmentation backbones}
\vspace{-0.5em}

\begin{table}[!b]
    \vspace{-0.5em}
    \centering
    \begin{tabu} to \columnwidth {@{}X[3,l]X[1,c]@{}}
    \toprule
    Model & mIoU(\%) \\
    \toprule
    \multicolumn{2}{l}{\textbf{\textit{Backbone:}} DeepLabV3+~\cite{chen2017deeplab}} \\
    Patch processing & 59.64 \\
    Downsampling & 52.01 \\
    MagNet & \textbf{61.99} \\
    \midrule
    \multicolumn{2}{l}{\textbf{\textit{Backbone:}} HRNetV2-W48 + OCR~\cite{ocr}} \\
    Patch processing & 54.30 \\
    Downsampling & 63.92 \\
    MagNet & \textbf{68.90} \\
    \bottomrule
    \end{tabu}
    \vskip -0.1in
    \caption{ {\small {\bf Results of using MagNet with two backbone networks}. MagNet can be used with different segmentation backbones, and improve their segmentation results. See \Tref{tab:cityscapes_result} for the results for using MagNet with HRNetV2-W18 backbone.}}
    \label{tab:cityscapes_backbones}
    \vspace{-0.5em}
\end{table}

We also tested the MagNet framework with two different segmentation backbone networks, and the results are shown in \Tref{tab:cityscapes_backbones}. In both cases, MagNet improved the segmentation results of the original networks significantly, from 2\% to 5\%. In this experiment, we used four scale levels: $256{\rightarrow}512{\rightarrow}1024{\rightarrow}2048$ and the number of refinement locations for each patch was $k=2^{16}$. 

\subsection{DeepGlobe}

\begin{table}[t!]
    \vspace{-0.25em}
    \centering
    \begin{tabu} to \columnwidth {@{}X[4,l]X[3,l]X[2.5,l]X[1.5,r]@{}}
    \toprule
    Model & mIoU(\%) & Mem.(MB) & Time(s)\\
    \toprule
    \multicolumn{3}{l}{\textbf{\textit{Downsampling}}} \\
    U-net\cite{unet} & 50.11 & 1813\\
    FCN-8s\cite{fcn} & 52.86 & 10569\\
    SegNet\cite{segnet} & 60.93 & 2645\\
    DeepLabv3+\cite{chen2017deeplab} & 63.50 & 1541\\
    FPN\cite{fpn} & 67.86 & 1247 & 0.01\\
    \midrule
    \multicolumn{3}{l}{\textbf{\textit{Patch processing}}} \\
    U-net\cite{unet} & 46.53 & 1813 \\
    FCN-8s\cite{fcn} & 62.43 & 10569\\
    SegNet\cite{segnet} & 68.40 & 2645\\
    DeepLabv3+\cite{chen2017deeplab} & 69.69 & 1541\\
    FPN\cite{fpn} & 70.98 & 1247 & 0.31\\
    \midrule
    DenseCRF\cite{densecrf} & 70.36~\smalldown{0.62} & 1247 & 39.68 \\
    DGF\cite{deepguidedfilter} & 70.38~\smalldown{0.6} & 1435~\smallup{188} & 0.25 \\
    GLNet\cite{chen2019collaborative} & 71.60~\smallup{0.62}  & 1865~\smallup{618} & 0.37 \\
    PointRend\cite{pointrend} & 71.78~\smallup{0.8} & 1593~\smallup{346} & 0.16  \\
    MagNet-Fast & 71.85~\smallup{0.87} & 1559~\smallup{312} & 0.29 \\
    MagNet & \textbf{72.96}~\smallup{1.98}  & 1559~\smallup{312} & 1.19 \\
    \bottomrule
    \end{tabu}
    \vskip -0.1in
    \caption{{\bf Segmentation results on the DeepGlobe dataset}. We used the same segmentation backbone (FPN) for all refinement methods in the last part.}
    \label{tab:deepglobe_result}
    \vspace{-0.75em}
\end{table}

DeepGlobe is a dataset of high-resolution satellite images. The dataset contains 803 images, annotated with seven landscape classes, including the \textit{unknown} class. Following the evaluation protocol of \cite{chen2019collaborative}, the \textit{unknown} class is ignored in mIoU calculation, so there are only six classes to consider. The size of the images is $2448{\times}2448$ pixels. We used the same train/validation/test split as reported in \cite{chen2019collaborative}, with 455, 207, and 142 images for training, validation, and testing, respectively.

The Feature Pyramid Network (FPN)~\cite{fpn} with Resnet-50 backbone was used as the segmentation network as in the previous work GLNet~\cite{chen2019collaborative}. We also used the same input size $508{\times}508$ as GLNet. We used three refinement stages with three scales $612{\rightarrow}1224{\rightarrow}2448$ and selected $2^{16}$ points for refinement at each scale. The results are shown in \Tref{tab:deepglobe_result}.
For MagNet-Fast, at each of the three scale levels, we selected the top three patches with the highest level of prediction uncertainty for refinement. PointRend~\cite{pointrend} was also trained with the same segmentation backbone and it achieved higher accuracy than GLNet. Both MagNet and MagNet-Fast outperformed other methods.

\subsection{Gleason}

\begin{table}[!t]
    \vspace{-0.25em}
    \centering
    \begin{tabu} to \columnwidth {@{}X[3.5,l]X[3,l]X[2.5,l]X[1.5,r]@{}}
    \toprule
    Model & mIoU(\%) & Mem.(MB) & Time(s) \\
    \midrule
    Experts & 65.48 & - & - \\
    Patching & 46.56 & 1903 & 2.42 \\
    Downsampling & 68.90 & 1903 & 0.02\\
    DenseCRF\cite{densecrf} & 69.46~\smallup{0.56} & 1903 & 141.79 \\
    DGF\cite{deepguidedfilter} & 68.91~\smallup{0.01} & 2223~\smallup{320} & 0.29  \\
    PointRend\cite{pointrend} & 68.97~\smallup{0.07} & 2655~\smallup{752} & 0.21 \\
    MagNet-Fast & 69.75~\smallup{0.85} & 2621~\smallup{718} & 0.33\\
    MagNet & \textbf{70.60}~\smallup{1.7} & 2621~\smallup{718} & 2.74\\
    \bottomrule
    \end{tabu}
    \vskip -0.1in
    \caption{{\bf Performance of MagNet and other frameworks on Gleason dataset with PSPNet~\cite{zhao2017pyramid} as the backbone}.}
    \label{tab:gleason_result}
    \vspace{-1.5em}
\end{table}

Gleason~\cite{gleason2019} is a medical dataset with histopathological images of prostate cancer. The task is to segment and grade lesions on ultra-high-resolution images. There are four classes in the dataset that need to be segmented: benign, Grade 3, Grade 4, and Grade 5. The dataset contains 244 images with a size of $5000{\times}5000$ pixels with segmentation labels provided by six clinical experts. The combined final label is based on majority voting. We randomly split the dataset into 195 training and 49 testing images.

PSPNet~\cite{zhao2017pyramid}, the highest-ranked method on the leaderboard for Gleason, was used as the segmentation network with the backbone Resnet-101. We used the input size of $512{\times}512$, and four refinement stages with four scales: $625{\rightarrow}1250{\rightarrow}2500{\rightarrow}5000$. MagNet-Fast was also run on four scales, but only on three patches with the highest level of prediction uncertainty at each scale. There are $2^{16}$ refinement points for each scale. The results of MagNet and MagNet-Fast, together with the result of the winning solution PSPNet and the mIoU agreement between medical experts, are shown in \Tref{tab:gleason_result}. MagNet was run with the PSPNet segmentation backbone, and it improved the performance of PSPNet by 1.7\%. 

 \vspace{-0.25em}
 
\section{Conclusions}
\vspace{-0.25em}
We have proposed MagNet, a multi-scale segmentation framework for high-resolution images. MagNet can generate high-resolution segmentation output without exploding the GPU memory usage by dividing input images into patches. To avoid the problem of being too global or local, patches of multiple scales are considered, from the coarsest to the finest levels. MagNet has multiple segmentation stages, where the output of one stage will be used as the input for the next stage, and the segmentation output will be progressively refined. We have demonstrated the benefits of MagNet on three challenging high-resolution image datasets, where MagNet outperforms the previous state-of-the-art methods by a margin of 1\% to 2\% in terms of mean Intersection over Union (mIoU).

{\small
    \setlength{\bibsep}{0pt}
    \bibliographystyle{plainnat}
    \bibliography{longstrings,egbib,pubs}
}

\end{document}